# DEEP ASYMMETRIC MIXTURE MODEL FOR UNSUPERVISED CELL SEGMENTATION

*Yang Nan, Guang Yang*

Department of Bioengineering, Imperial College London
Imperial-X, Imperial College London

## ABSTRACT

Automated cell segmentation has become increasingly crucial for disease diagnosis and drug discovery, as manual delineation is excessively laborious and subjective. To address this issue with limited manual annotation, researchers have developed semi/unsupervised segmentation approaches. Among these approaches, Deep Gaussian mixture model plays a vital role due to its capacity to facilitate complex data distributions. However, these models assume that the data follows symmetric normal distributions, which is inapplicable for data that is asymmetrically distributed. These models also obstacles weak generalization capacity and are sensitive to outliers. To address these issues, this paper presents a novel asymmetric mixture model for unsupervised cell segmentation. This asymmetric mixture model is built by aggregating certain multivariate Gaussian mixture models with log-likelihood and self-supervised based optimization functions. The proposed asymmetric mixture model outperforms (nearly 2-30% gain in dice coefficient, $p<0.05$) the existing state-of-the-art unsupervised models on cell segmentation including the segment anything.

## 1. INTRODUCTION

Cell segmentation provides useful information for disease diagnosis and quantitative analysis in biomedical research. However, pathologists can barely count millions of cells from whole slide images. To address this issue, researchers have developed semi/un-supervised approaches, including conventional approaches (e.g., marker-controlled watershed [1], K/C Means [2], and region based approaches [3]). Unfortunately, these methods were found to be inaccurate and have low generalisation abilities. For instance, the performance of these models dramatically fluctuates on multicentre datasets, due to the variations among images.

With the inspiring results of deep learning, researchers have combined deep neural networks with conventional approaches. Among these methods, deep clustering methods have received much attention due to their feasibility and scalability. Deep clustering approaches usually include feature extraction, dimensionality reduction, and clustering optimisation functions [4, 5]. However, study have shown

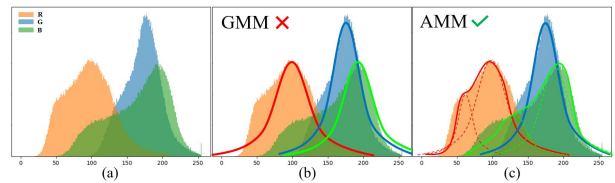

**Fig. 1.** (a) Histogram of the cell regions in a pathological image, and the results given by (b) GMM and (c) AMM. AMM is a mixture (red, green, and blue solid lines in (c)) of certain Gaussian distributions (dotted lines in (c)) and can estimate the asymmetric distribution in real applications.

that these deep clustering methods could not perform well on pathological data [6].

Efforts have also been made in mixture model based clustering, which partition datasets into different clusters by assuming the data is composed of a mixture of probability distributions. Specifically, each cluster accounts for a probability distribution, and the overall data distribution is generated by the mixture of different probability distributions. One common example of mixture model based clustering is the Gaussian mixture model (GMM), which assumes that each cluster follows a Gaussian distribution, as Fig. 1 (b). GMM estimates the mean and variance of each component and the probabilities of each data point belonging to different components. Meanwhile, the parameters of GMM can be intuitively estimated by introducing expectation maximisation, without complex computational procedures. Based on these advances, researchers developed deep Gaussian mixture models for the unsupervised learning task. An early attempt was in 2014 when Oord et al. [7] built a GMM after the top layers of a classification neural network. In addition, Zong et al. [8] combined GMM with an auto-encoder decoder to recognize anomalies. Zanjani et al. [9] proposed a deep Gaussian mixture model to segment pathological images for colour normalisation. Nevertheless, existing study [6] have shown that GMM is sensitive to outliers and could lead to degenerative issues such as class domination and poor reproducibility. To address this issue, Nan et al. [6] presented a deep-constrained Gaussian mixture model for unsupervised segmentation. However, the current studies did not consider the case when the data is asymmetrically distributed. In the real world, regions of interest are usually distributed asymmetrically, resulting in an inaccurate estimation by deep mixture models. For instance, Fig. 1 (a) plots the distribution of cells in a pathological image, while all distributions of R, G, and B channels appear to be asymmetric. As a result, one can hardly use GMM to estimate these kinds of distributions.

This study presents a novel deep asymmetric mixture model (DAMM) for unsupervised segmentation. The proposed method builds asymmetric mixture models (AMM) by introducing hierarchical Gaussian mixture models. Specifically, each component of the proposed DAMM is generated from a multivariate GMM to estimate asymmetric distribution. Additionally, a self-supervised learning scheme is introduced to help the model learn rotate-invariant features better in order to improve the model performance. Comprehensive experiments have been performed based on a large open-access dataset Lizard [10]. Inspired by [6], repeated experiments have been conducted to evaluate the stability and robustness of all comparison models.

The main contribution of the proposed study is summarized as follows:
- We propose a fully unsupervised method for cell segmentation, addressing the challenge of asymmetric data distributions.
- A rotation-invariant optimisation module is integrated into the loglikelihood maximisation protocol to improve the model performance using gradient descent.
- We conduct comprehensive experiments that incorporate recent SOTA approaches such as segment anything (SAM) [11], one of the foundation models for segmentation.

## 2. METHOD

**Asymmetric mixture model**

The proposed asymmetric mixture model is defined by introducing multi-level mixture models. Given the task of estimating the posterior probability of Assume $f(x_i|\Theta)$ as the probability density function (PDF) of the proposed multivariate asymmetric distribution

$$f(x_i|\Theta) = \sum_{j=1}^{K} \pi_j \, p(x_i|K), \quad (1)$$

where $\pi_j$ is the mixture weight of the proposed asymmetric mixture model, $p(x_i|K)$ is the $K$-th component for the AMM, which can be any type of mixture models. In this study, we use the Gaussian mixture model to constitute the component,

$$p(x_i|K) = \sum_{j=1}^{K} \pi_j \sum_{m}^{M} \alpha_{j,m} \phi(x_i|\mu_{j,m}, \Sigma_{j,m}). \quad (2)$$

Assume $D$ is the dimension of data $x \in R^{N \times D}$ (N is the number of samples and D is the data dimension), $|\Sigma_{j,m}|$ is the determinant of the matrix $\Sigma_{j,m}$, $\mu_{j,m}$ is the mean and $\alpha_{j,m}$ is the mixture weight of sub-mixture models, the probability density function of multivariate Gaussian distribution $\phi$ can be given from

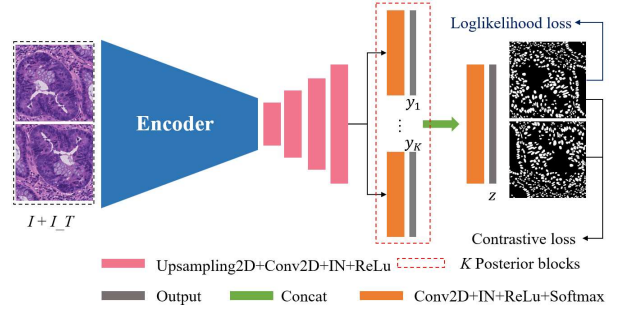

**Fig. 2.** Overview of the proposed DAMM, including an encoder, a decoder, K posterior blocks to generate gaussian mixture distribution $p(x_i|C_j)$, and a hybrid optimization function which consists of a loglikelihood loss and a contrastive loss. '*I*' and '*I_T*' represent the original image and transferred image (for rotation-invariant feature learning), respectively.

$$\phi(x_i|\mu_{j,m}, \Sigma_{j,m}) = \frac{1}{2\pi^{\frac{D}{2}}\sqrt{|\Sigma_{j,m}|}} e^{-\frac{1}{2}(x_i-\mu_{j,m})^T \Sigma_{j,m}^{-1}(x_i-\mu_{j,m})}. \quad (3)$$

Based on Eq. (2) and Eq. (3), the log-likelihood function can be given by

$$L(\Theta) = \sum_{i=i}^{N} \log \left[ \sum_{j}^{K} \pi_j \sum_{m}^{M} \alpha_{j,m} \phi(x_i|\mu_{j,m}, \Sigma_{j,m}) \right] \quad (4)$$

where $\Theta = [\pi_j, \alpha_{j,m}, \mu_{j,m}, \Sigma_{j,m}]$. Therefore, the parameter estimation of the asymmetric mixture model can be given by maximizing the loglikelihood function $L(\Theta)$. Assume $z_{i,j}$ as the posterior probability of the asymmetric mixture model, $y_{i,j,m}$ as the posterior probability of the Gaussian mixture model $p(x_i|C_j)$, according to Jensen's inequality, the loglikelihood function can be further calculated through

$$L(\Theta) = \sum_{i=1}^{N} \sum_{j=1}^{K} z_{i,j} [\log \pi_j + \sum_{m}^{M} y_{i,j,m} (\log \alpha_{j,m} + \mathbb{E}_\Theta (\log \phi(x_i|\mu_{j,m}, \Sigma_{j,m})))] \quad (5)$$

**Deep Asymmetric mixture model with contrastive learning**

The architecture of the proposed deep asymmetric mixture model mainly includes an encoder for feature extraction and a decoder for mask generation (shown in Fig. 2). To prevent overfitting when performing loglikelihood maximization through a deep learning scheme, the encoder $E$ is built followed by MobileNet-V3 [12], a light and efficient architecture. To acquire the pseudo posterior probability $z_{i,j}$ and $y_{i,j,m}$, $K$ posterior blocks are proposed after the last layer of the decoder $\mathcal{D}$. These posterior blocks aim to generate the pseudo posterior probability $y_{i,j,m}$ of the $j$-th Gaussian mixture model $p(x_i|C_j)$. In addition, the pseudo posterior probability $z_{i,j}$ of the asymmetric mixture model is acquired by conducting basic operations (convolution layer, instance normalization layer, and ReLU) on the concatenation of all $y_{i,j,m}$.

Given an RGB input image $I \in R^{W \times H \times 3}$ and its flattening format $I' \in R^{(W*H) \times 3}$, parameters of the proposed DAMM can be updated by making the partial derivative over each parameter from Eq. (5),

$$\{z_{i,j}^{(t+1)}, y_{i,j,m}^{(t+1)}\} = \mathcal{D}(E(I)), \quad (6)$$

$$\mu_{j,m}^{(t+1)} = \frac{\sum_{i=1}^{N} z_{i,j}^{(t+1)} y_{i,j,m}^{(t+1)} I'_i}{\sum_{i=1}^{N} z_{i,j}^{(t+1)} y_{i,j,m}^{(t+1)}}, \quad (7)$$

$$\Sigma_{j,m}^{(t+1)} = \frac{\sum_{i=1}^{N} z_{i,j}^{(t+1)} y_{i,j,m}^{(t+1)} (I'_i - \mu_{j,m}^{(t+1)})(I'_i - \mu_{j,m}^{(t+1)})^T}{\sum_{i=1}^{N} z_{i,j}^{(t+1)} y_{i,j,m}^{(t+1)}}, \quad (8)$$

$$\pi_j^{(t+1)} = \frac{1}{N} \sum_{i=1}^{N} z_{i,j}^{(t+1)}, \quad (9)$$

$$\alpha_{j,m}^{(t+1)} = \frac{\sum_{i=1}^{N} z_{i,j}^{(t+1)} y_{i,j,m}^{(t+1)}}{\sum_{i=1}^{N} z_{i,j}^{(t+1)} \sum_{m=1}^{M} y_{i,j,m}^{(t+1)}}. \quad (10)$$

In addition to minimizing the negative loglikelihood $L(\Theta')$, a self-supervised learning optimization function is presented to enhance the efficiency of feature extraction by emphasizing the translation invariance of image predictions. For each training iteration, the training batch size is set to 2, with the raw image $I$ and its transformed image $I_T$, where $I_T = f_T(I)$. In this study, we applied randomized vertical/horizontal flips for the contrastive transform $f_T$. Therefore, the contrastive loss can be given through

$$L_C = |f_T(\mathcal{D}(E(I))) - \mathcal{D}(E(I_T))|. \quad (11)$$

Overall, the optimization function for the proposed DAMM can be calculated as

$$L = -L(\Theta') + \beta L_C, \quad (12)$$

where $\beta$ is the weight of contrastive loss, $\omega_E, \omega_\mathcal{D}$ are the trainable parameters in encoder $E$ and decoder $\mathcal{D}$, and $\Theta' = [\pi_j, \alpha_{j,m}, \mu_{j,m}, \Sigma_{j,m}, \omega_E, \omega_\mathcal{D}]$. During the training, the model can be saved regarding the minimum loss values or the convergence of log-likelihood. Alternatively, using a fixed image with ground truth can improve the efficiency of model selection by calculating its overlapped measurements in each validation process.

## 3. EXPERIMENTS

**Dataset and evaluation metrics**

Lizard dataset [10] is an open dataset collected from H&E-stained colon tissue (20x objective magnification). Six types of cells were annotated including epithelial cells, connective tissue cells, lymphocytes, plasma cells, neutrophils and eosinophils, with the total number of 495,179. It was collected from six data resources, indicating sufficient variability and diversity. In this study, we followed the ratio of 7:2:1 to split all 238 images randomly into training and validation sets. The sizes of the original 238 images range from 500×500 pixels to 2019×2175 pixels, while we cropped $512 \times 512$ patches by sliding window and resizing when necessary (images with sizes less than 512×512) operations. In particular, there were 1203 images for training and 445 images for testing in this study.

The evaluation metrics include Dice coefficient scores, with Wilcoxon signed-rank test to assess the significant differences between different approaches (we consider P<0.05 as significant, and P<0.001 as highly significant). Meanwhile, in addition to SAM, repeated experiments (10 times) were conducted to assess the robustness and stability of all comparison methods.

**Training strategies and experimental settings**

All the trainable parameters of our DAMM were initialized using Xavier initialization. Randomized horizontal flips, vertical flips, hue adjustment (factor=0.12), saturation adjustments (factor=0.5) and brightness adjustments (factor=0.15) were conducted for online data augmentation. All the models were developed on: Ubuntu 20.04.2 LTS with CPU i9-10980XE, NVIDIA RTX 3090 GPU, 128GB memory; CUDA 11.6.1; Python 3.8.0 with Pytorch 1.13.1.

The comparison approaches include classic unsupervised approaches including minibatch K-Means (mKMeans) and GMM, and recent deep unsupervised segmentation approaches including IIC [13], DFC [4], DCGMM [9], DCGN [6], and SAM [11]. All comparison experiments were trained with a batch size of 2, using Adam optimiser with an initial learning rate of 1e-4 and exponential learning rate reduce protocol (gamma=0.97). The β used in Eq. (12) was empirically set as 0.5 in this study. For a fair comparison, all the (deep learning based) comparison experiments used the same segmentation architecture shown in Fig. 2 (without posterior blocks) and the same training strategy. M is set to 2 and K is set to 3 for all comparisons. Details of the MobileNet V3 used (as DAMM backbones) in this study can be further found at https://github.com/kuan-wang/pytorch-mobilenet-v3.

## 4. RESULTS AND DISCUSSION

The results of 10 repeated experiments are summarized in Table 1. For each round, models were trained from scratch on (the same) 1,203 images and tested on (the same) 445 images. It is of note that, all methods were trained fully unsupervised, and ground truth was only used during metric calculation without information leakage. The time costs for all approaches were calculated during the inference stage and the Dice scores were statistically analyzed by performing the Wilcoxon signed-rank test.

From Table. 1, DAMM achieves the best average performance of 0.6295 Dice coefficient score among all unsupervised learning approaches, with highly significant differences of p<0.001. Meanwhile, in 10 rounds of repeated evaluations, DAMM also presents high stability (low standard deviation) and accuracy (10-round Dice coefficient score of 0.6237). In addition to DAMM, DCGN presented highly competitive results, with 0.6112 Dice scores and similar inference time. Surprisingly, previous approaches such as DFC and IIC both failed to produce satisfactory results (0.3544 and 0.3692 Dice scores, respectively), which may be due to the complexity and high intra-class variation of pathological patterns. Interestingly, GMM presented better than DCGMM, and achieved the best stability among all comparison studies with a standard deviation of 0.0015. It is of note that the large foundation model SAM did not perform well on cell segmentation with an dice score of 0.4462.

**Table 1.** Comprehensive performance of unsupervised segmentation

| Methods | Dice-Best | Dice-Average | Time cost (s) |
| --- | --- | --- | --- |
| mKMeans | 0.4559† | 0.4503±0.0037† | 0.016† |
| GMM | 0.5095† | 0.5070±0.0015† | 0.049† |
| DFC | 0.3544† | 0.3237±0.1635† | 4.187† |
| IIC | 0.3692† | 0.3449±0.0111† | 0.084† |
| DCGMM | 0.4417† | 0.4242±0.0175† | 0.034 |
| DCGN | 0.6112† | 0.5995±0.0107† | 0.033 |
| SAM | | 0.4462±0.1664† | 4.51† |
| OURS | 0.6295 | 0.6237±0.0051 | 0.039 |

† indicates highly significant difference compared with DAMM, with Wilcoxon signed-rank test P<0.001. 'Dice-Best' indicates the average performance of the best round of 10 repeated experiments, and 'Dice-Average' refers to the average performance of 10 rounds of repeated experiments (shown as mean ± std).

In addition to quantitative performance, visualisation results are given in Fig. 3. It can be found that both GMM and DCGMM suffered heavy false positive (FP) predictions and all methods have cell adhesion issues. mKMeans has less FP predictions while it suffered from false negatives (FN). Interestingly, the large foundation model SAM could well address the adhesion issue, while it failed to produce semantic predictions. For instance, SAM failed to predict cells within the glands when glands were segmented as instances. DCGN appears to have a similar performance to DAMM, while it has more false negative predictions. The proposed DAMM achieved the best performance among all comparison experiments, with considerable accuracy and robustness.

Furthermore, a typical example distribution of predictions given by DAMM has been plotted in Fig. 4. It can be found that the pixel values of the ground truth (cells) are distributed asymmetrically, while the proposed DAMM can better model the real distributions. Although DCGN appears similar performance to DAMM in some cases, it cannot well model asymmetric distribution as in Fig. 4. This might be because the 'mixture' of "mixture models" provides more intra-class variation tolerance. However, limitations also exist for the current DAMM, since the number of subcomponents still requires manual settings (K is set to 2 in this study), which may lead to overfitting issues during the training progress.

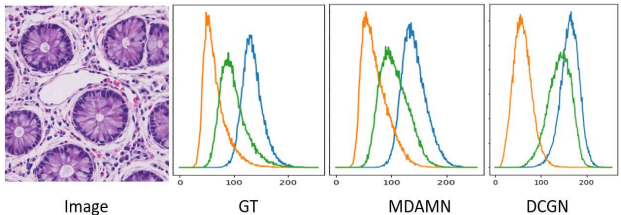

**Fig. 4.** Distribution of ground truth annotation and prediction given by DAMM and DCGN.

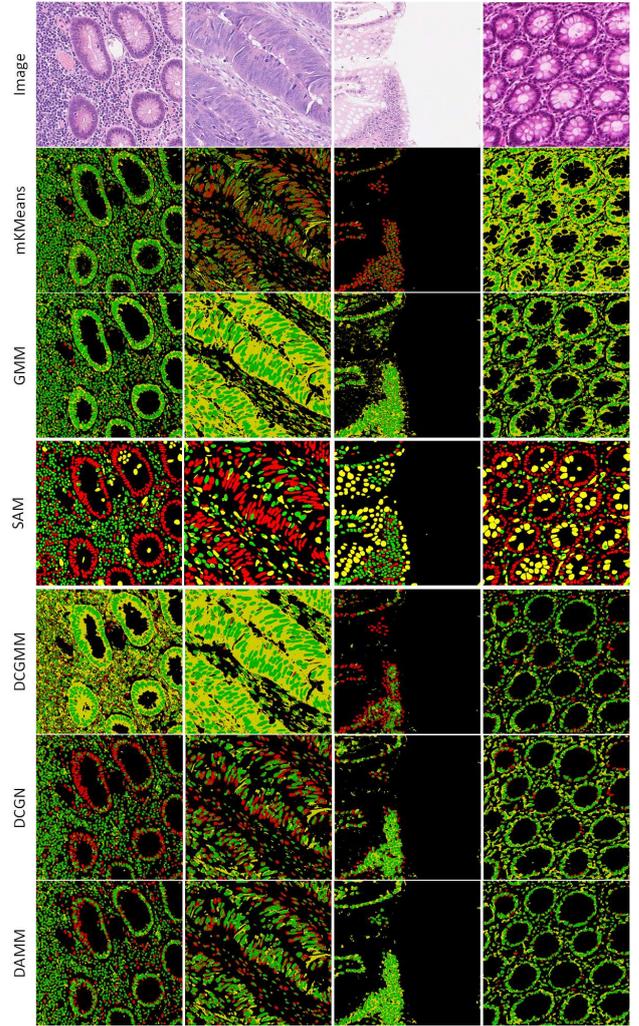

**Fig. 3.** Visualization results of unsupervised approaches (GMM, DCGN, DAMM, etc.) and large foundation model SAM. The green, yellow, and red colours refer to true positive, false positive and false negative predictions, respectively.

## 5. CONCLUSION

This paper presents a novel yet effective approach for unsupervised cell segmentation tasks, employing an asymmetric mixture model and a contrastive learning scheme. The proposed method achieved state-of-the-art performance on an open-access dataset, compared with the existing studies. We can envision a better performance by incorporating more contrastive learning strategies or vision Transformers architectures in our future work.